\documentclass[sigconf]{acmart}
\AtBeginDocument{%
  }

\setcopyright{acmlicensed}
\copyrightyear{2025}
\acmYear{2025}
\acmDOI{XXXXXXX.XXXXXXX}
\acmConference[UKAIR '25]{UK AI Research Symposium}{September 08--09,
  2025}{Newcastle, UK}
\acmISBN{978-1-4503-XXXX-X/2018/06}

\usepackage{balance}

\begin{document}

\title{Safe and Socially Aware Multi-Robot Coordination in Multi-Human Social Care Settings}


\author{Ayodeji O. Abioye}
\email{ayodeji.abioye@open.ac.uk}
\orcid{0000-0003-4637-3278}
\affiliation{%
  \institution{School of Computing and Communications, The Open University}
  \city{Milton Keynes}
  \country{UK}
}

\author{Jayati Deshmukh}
\email{j.deshmukh@soton.ac.uk}
\orcid{}
\affiliation{%
  \institution{School of Electronics and Computer Science, University of Southampton}
  \city{Southampton}
  \country{UK}
}

\author{Athina Georgara}
\email{a.georgara@soton.ac.uk}
\orcid{}
\affiliation{%
  \institution{School of Electronics and Computer Science, University of Southampton}
  \city{Southampton}
  \country{UK}
}

\author{Dominic Price}
\email{dominic.price@nottingham.ac.uk}
\orcid{}
\affiliation{%
  \institution{School of Computer Science, University of Nottingham}
  \city{Nottingham}
  \country{UK}
}

\author{Tuyen Nguyen}
\email{tuyen.nguyen@soton.ac.uk}
\orcid{}
\affiliation{%
  \institution{School of Electronics and Computer Science, University of Southampton}
  \city{Southampton}
  \country{UK}
}

\author{Aleksandra Landowska}
\email{aleksandra.landowska@nottingham.ac.uk}
\orcid{}
\affiliation{%
  \institution{School of Computer Science, University of Nottingham}
  \city{Nottingham}
  \country{UK}
}

\author{Amel Bennaceur}
\email{amel.bennaceur@open.ac.uk}
\orcid{}
\affiliation{%
  \institution{School of Computing and Communications, The Open University}
  \city{Milton Keynes}
  \country{UK}
}

\author{Joel E. Fischer}
\email{joel.fischer@nottingham.ac.uk}
\orcid{0000-0001-8878-2454}
\affiliation{%
  \institution{School of Computer Science, University of Nottingham}
  \city{Nottingham}
  \country{UK}
}

\author{Sarvapali D. Ramchurn}
\email{sdr1@soton.ac.uk}
\orcid{0000-0001-9686-4302}
\affiliation{%
  \institution{School of Electronics and Computer Science, University of Southampton}
  \city{Southampton}
  \country{UK}
}

\renewcommand{\shortauthors}{Abioye et al.}

\begin{abstract}
  This research investigates strategies for multi-robot coordination in multi-human environments. It proposes a multi-objective learning-based coordination approach to addressing the problem of path planning, navigation, task scheduling, task allocation, and human-robot interaction in multi-human multi-robot (MHMR) settings.
\end{abstract}



\received{20 February 2007}
\received[revised]{12 March 2009}
\received[accepted]{5 June 2009}

\maketitle



    
\section{Introduction}
Robots in co-located human spaces may be required to interact with humans or move safely around them. A coordination problem arises when multiple robots share the same human space. For example, which robot gives way between a cleaning robot and a delivery trolley robot on a narrow passage, or which robot interacts with the human between a social robot and an aerial robot relaying critical information, or which robot does a delivery between an aerial robot and a delivery trolley robot when both are equally capable of completing the delivery. This results in a multi-objective coordination problem of navigation, interaction, and task allocation, a combination required to extend the capability and utility of functional robots in co-located human spaces 
\cite{shiokawa_beyond_2025,kramer_i_2025,abioye2022_msvg}.

A challenge for co-located robot navigation is the limited understanding of human behaviours and prediction of future actions \cite{mavrogiannis_core_2023,moller_survey_2021,cheng_autonomous_2018}. Several algorithms have been proposed to address robot navigation in human spaces \cite{cui_path_2025,tolani_visual_2021, Perez-darpino_robot_2021,cosgun_anticipatory_2016}, often modelling the human as a dynamic obstacle. Some learning-based approaches include \cite{dugas_navrep_2021,liu_robot_2020}. Human-aware algorithms \cite{teja_singamaneni_human-aware_2021,guzzi_human-friendly_2013} focus on the individual human-robot encounter. Social-aware algorithms \cite{hoang_socially_2023,kivrak_social_2021,truong_toward_2017} take into account the presence of multiple humans. Some other approach involves mapping safe zones around the co-located humans \cite{abioye2024_mapping_safe_zones,svenstrup_trajectory_2010}. However, there are very limited works exploring multi-robot coordination in multi-human environments \cite{wang_initial_2024,heuer_benchmarking_2024}.

In this research, we seek to address the multi-robot coordination challenge by going beyond ``obstacle avoidance'' or ``shortest distance to target'', modelling the social context, taking into account the historical and environmental context, combined with the physical and physiological sensor data in path-planning, navigation, task scheduling, task allocation, human interaction, and other action execution.



\begin{figure*}[!htb]
  \centering
  \framebox{\includegraphics[width=0.84\textwidth]{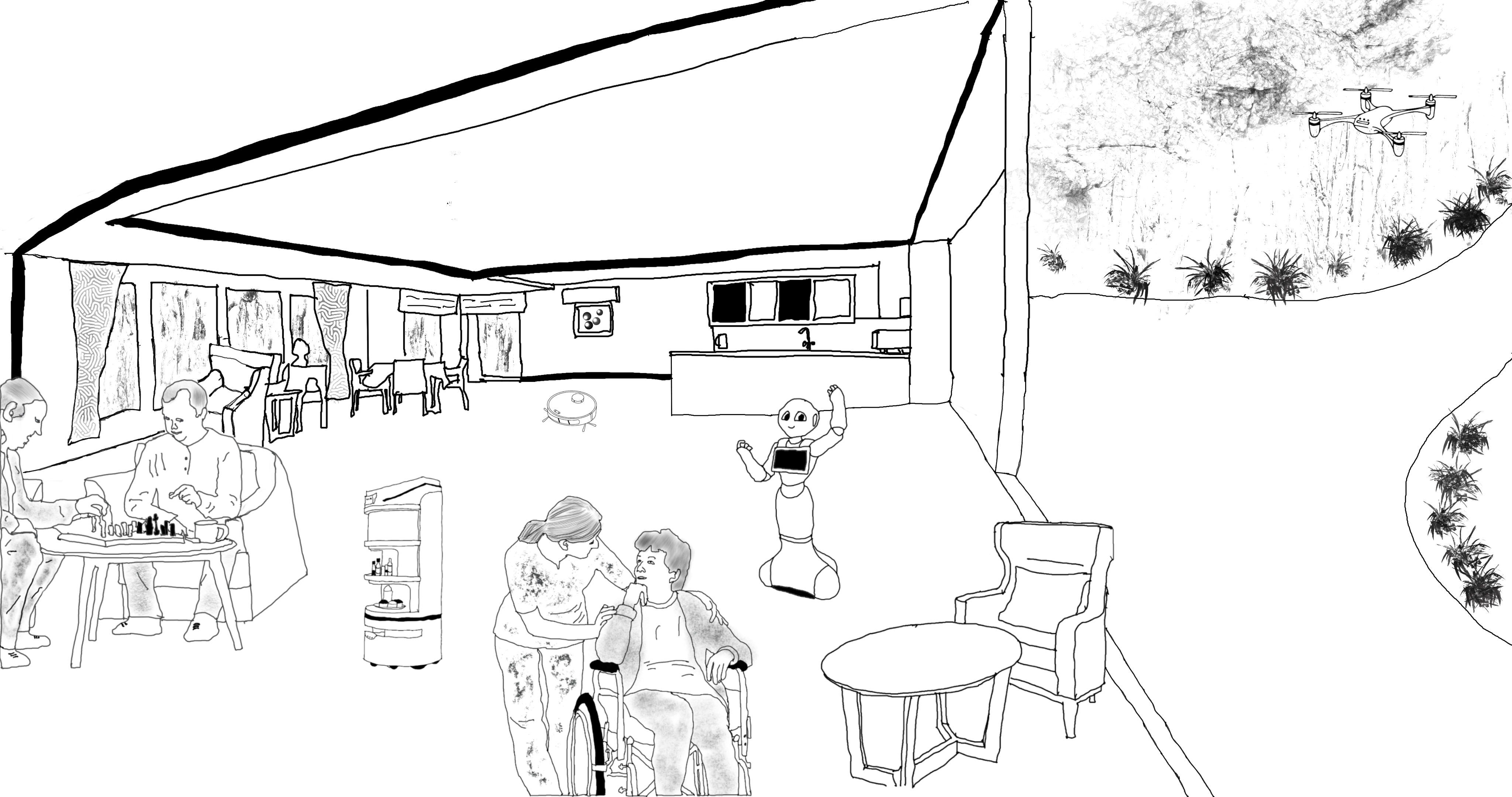}}
  \caption{Conceptualising multiple robots in a social care setting consisting of three residents, one carer, an aerial robot, a cleaning robot, a delivery trolley robot, and a social robot.}
  \Description{A three-dimensional sketch of a care home communal area opening into a garden. The image consists of two residents playing chess, one resident in a wheelchair being attended by a care worker, a vacuum cleaning robot cleaning, a social robot roaming about, a delivery trolley robot located next to the residents playing chess, and a drone patrolling the garden areas.}
  \label{fig:social_care_setting}
\end{figure*}

\begin{table}
  \caption{Multi-human multi-robot (MHMR) distribution}
  \label{tab:mhmr_distribution}
  \begin{tabular}{ll}
    \toprule
    Agent & Roles\\
    \midrule
    Residents & Humans needing care and support\\
    Carers & Humans supporting residents\\
    Aerial Robot & Patrol - activity and hazard tracking\\
    Cleaning Robot & Corridors and communal area Cleaning\\
    Trolley Robot & Medication, drinks, supplies delivery\\ 
    Social Robot & Entertainment\\
  \bottomrule
\end{tabular}
\end{table}

\section{Methodology}
The scenario consists of a medium-sized residential care facility for the elderly as conceptualised in Figure \ref{fig:social_care_setting}, consisting of four robots with roles described in Table \ref{tab:mhmr_distribution}. All residents, carers, and robots wear or carry an ultra-wideband tag to make them easier to locate. Robots use facial recognition to identify residents.
The aerial robot is used for monitoring the outdoor areas (garden/perimeter). 
The action set of the aerial robot is given as $A_a = \{$patrol, return to base, hover, descend, alert$\}$. The first problem is to determine an optimal path-planning algorithm for navigating safely around co-located humans, especially when flying at human height levels under trees and building archways during the grounds patrol.
The cleaning robot is used for cleaning flat indoor surfaces. 
Its action set are $A_c = \{$move forward, rotate, clean, stop, dock$\}$. The second problem is to autonomously determine when to clean and how to clean safely, for example, immediately after a spill to avoid an accident.
The social robot supports residents by keeping them engaged conversationally, facilitating exercise activities, and participating in games such as trivia. 
Its action set $A_s = \{ \text{speak}, \text{wave}, \text{play music}, \text{ask question}, \text{call carer} \}$ with a policy-driven behaviour model, $\pi: O_i \rightarrow A_s$, where \( O_i \) are observed inputs (speech, emotion cues from residents). The third problem is about gathering care-specific behavioural data that can be used to train the robot's model. How can the robot interpret relevant resident or carer actions, combining this with data from other robots, to determine the socially acceptable response?
The trolley robot supports carers by delivering medications, drinks, snacks, and other items. 
Its action set $A_t = \{$navigate to target location, find recipient, dispense item, wait, return to dock$\}$. The fourth problem is about how to find an item recipient since they are dynamic. How can it leverage information from other robots to locate an item's recipient while navigating in a safe and socially compliant manner?

\section{Discussion and Further Works}
Coordinating multiple robots in multiple human settings is not a simple task. Technical challenges include physical robot design, real-world sensing, path planning, navigation, task scheduling, human activity recognition, human-robot interaction, robot-robot interaction, hazard perception, and information sharing. Knowing the positions of humans behind walls or doors combined with learned human behaviour means the robot path planning algorithm can include the actions of unseen humans, such as moving towards a door that opens into the robot's path. The robot can start slowing down before its physical sensors detect the human, resulting in safer and socially compliant robot behaviour.



\begin{acks}
This research was conducted as part of the UKRI Responsible AI (EP/Y009800/1) research grant's project on ``Embodied AI in Social Spaces: Responsible and Adaptive Robots in Complex Settings''.
\end{acks}

\balance

\bibliographystyle{ACM-Reference-Format}
\bibliography{ref}

\end{document}